\documentclass[journal]{IEEEtran}

\usepackage{cite}
\usepackage{amsmath,amssymb,amsfonts}
\usepackage{algorithm,algorithmic}
\usepackage{graphicx}

\usepackage{textcomp}

\usepackage{booktabs}
\usepackage{multirow}
\usepackage{makecell}
\usepackage{array}
\usepackage{colortbl}
\usepackage{rotating}
\usepackage{tabularx}
\usepackage{bm}

\usepackage{subcaption}
\usepackage{float}
\usepackage{placeins}
\usepackage{stfloats}
\usepackage{balance}

\def\BibTeX{{\rm B\kern-.05em{\sc i\kern-.025em b}\kern-.08em
    T\kern-.1667em\lower.7ex\hbox{E}\kern-.125emX}}

\begin{document}

\title{Structural-Functional Brain Connectivity Generation via
Multimodal Hypergraph-based Flow Matching}

\author{Chyong~Yi~Poh, Hwa~Hui~Tew, Junn~Yong~Loo, Rapha\"{e}l C.-W. Phan, Fuad~Noman,
Pew-Thian~Yap, and Chee-Ming~Ting
\thanks{This work has been submitted to the IEEE for possible publication. Copyright may be transferred without notice, after which this version may no longer be accessible.}
\thanks{Chyong Yi Poh, Hwa Hui Tew,  Rapha\"{e}l C.-W. Phan, Fuad Noman, and Chee-Ming Ting are
with the School of Information Technology, Monash University Malaysia
(e-mail: chyong.poh@monash.edu; hwa.tew@monash.edu; raphael.phan@monash.edu; fuad.noman@monash.edu;
 ting.cheeming@monash.edu).}
\thanks{Pew-Thian Yap is with the Department of Radiology, University
of North Carolina at Chapel Hill, Chapel Hill, NC, USA
(e-mail: ptyap@med.unc.edu).}
\thanks{Junn Yong Loo is with the School of Electrical and Electronic Engineering, Nanyang
Technological University, Singapore
(e-mail: junnyong.loo@ntu.edu.sg).}
\thanks{Corresponding author: Chee-Ming Ting
(e-mail: ting.cheeming@monash.edu).}}

\maketitle
\begin{abstract}
Structural connectivity~(SC) and functional connectivity~(FC) provide
complementary information on interactions between brain regions and are
widely used in neuroimaging studies of neuropsychiatric disorders.
Generative modelling can alleviate the scarcity of large-scale paired
SC--FC data, but existing approaches typically use pairwise graphs that
capture only dyadic interactions and often generate SC and FC
independently, limiting preservation of higher-order structure-function
relationships. We propose a \textit{Multimodal Hypergraph Flow Matching}
(MHG-FM) framework for joint SC--FC connectivity generation and
cross-modal translation. MHG-FM constructs modality-specific
hypergraphs, learns higher-order representations with Hypergraph Neural
Network~(HGNN) encoders, and performs bidirectional cross-modal fusion
using Dual Cross-Attention~(DCA). A variational autoencoder maps the
fused representations to a compact latent space, where conditional flow
matching enables connectivity synthesis and multimodal translation via
latent transport. Experiments on the Human Connectome Project Young
Adult~(HCP-YA) dataset show that MHG-FM outperforms several
state-of-the-art baselines in reconstruction quality, topology
preservation, distributional similarity, and SC--FC coupling, while
achieving approximately $8\times$ faster sampling than a matched
diffusion backbone.
\end{abstract}

\begin{IEEEkeywords}
Brain connectivity networks, Hypergraphs, Flow matching, Functional MRI, Diffusion MRI, Structural-functional coupling.
\end{IEEEkeywords}

\section{INTRODUCTION}
\label{sec:intro}

Characterizing brain connectivity provides insights into the organizational principles of the human brain.
Structural connectivity (SC) captures the anatomical organization of white matter pathways and provides the structural scaffold for temporal interactions across brain regions, thereby shaping the network topology of functional connectivity (FC)\cite{sporns2011,bullmore2009,friston2011}. The interaction between SC and FC provides complementary perspectives on brain organisation and disruption in brain networks is associated with disorders such as Major Depressive Disorder (MDD), Alzheimer's Disease (AD), and Autism Spectrum Disorder (ASD) \cite{zeng2012,ayyash2021}. Moreover, structure-function coupling is increasingly recognized as an important characteristic of brain organisation that evolves across the lifespan \cite{honey2009predicting,suarez2020linking,zimmermann2018,taylor2026}. Jointly modeling SC and FC is therefore important for characterising the interdependence between brain structure and function. Nevertheless, the simultaneous acquisition of paired multimodal neuroimaging data (SC and FC) remains challenging, costly, and time-consuming, often resulting in limited sample sizes~\cite{marek2022reproducible}. 

Generative modeling offers a promising strategy for this data scarcity. Early approaches used graph generative models such as GraphVAE \cite{simonovsky2018}, NetGAN \cite{bojchevski2018}, and GraphRNN \cite{you2018graphrnn}, but often faced challenges in training stability and optimization. More recently, graph diffusion frameworks such as DiGress \cite{vignac2023digress} have emerged as an alternative, improving the realistic synthesis of graph-structured data. Building on these methods, several neuroimaging studies have applied graph-based generative models for brain connectivity generation and cross-modal translation \cite{tan2025grspd,tan2025sfcgan,bessadok2021,zong2023brainnetdiff}. Although effective, these methods represent brain networks as simple pairwise graphs and cannot explicitly model higher-order interactions among multiple brain regions \cite{battiston2020}. Furthermore, current approaches typically generate SC and FC independently and focus on one-way translation, limiting their ability to explicitly model and preserve the SC-FC coupling mechanism.

To address these limitations, several works have explored hypergraph learning, in which a single hyperedge can connect multiple brain regions
simultaneously. For instance, hypergraph neural networks have shown strong performance in functional brain network analysis and disorder
classification~\cite{feng2019hgnn,bai2021,ji2022fchat}, including
recent extensions to multi-site, multi-atlas representation
learning~\cite{wang2024multiview} and multimodal hypergraph representations for disease diagnosis~\cite{xi2023hypergraph}. 
However, existing hypergraph studies are predominantly discriminative, with relatively limited exploration of generative modelling.  Diffusion-based hypergraph generation such as HYGENE~\cite{gailhard2025hygene} has achieved high-fidelity generative performance, but is not designed for multimodal generation nor coupling preservation. Moreover, generative efficiency remains a concern for diffusion models which

typically require costly iterative denoising ~\cite{ho2020ddpm,song2021score,zong2023brainnetdiff}. In contrast, Flow Matching (FM)~\cite{lipman2023} learns a continuous-time velocity field that transports samples from the source to the target distribution, offering comparable quality at lower sampling cost. Nevertheless, its application to multimodal hypergraph-based
brain connectivity generation remains unexplored.

To address the above-mentioned limitations, we propose Multimodal Hypergraph Flow Matching (MHG-FM), a framework for joint SC--FC connectome generation and cross-modal translation. MHG-FM constructs modality-specific SC and FC hypergraphs, encodes higher-order representations via Hypergraph Neural Network (HGNN) encoders, fuses them through a Dual Cross-Attention (DCA) \cite{luo2026scdiffusionx} mechanism, and optimises a variational latent representation via conditional flow matching to learn continuous transport trajectories for synthesis and translation.
Our contributions in this paper are summarized as follows:
\begin{enumerate}
\item We propose MHG-FM, a unified multimodal hypergraph flow-matching
framework for joint structural-functional (SC--FC) brain connectivity
generation and bidirectional cross-modal translation. It integrates
modality-specific hypergraph encoders, cross-modal latent fusion, and
conditional flow matching in a shared variational latent space to model
higher-order SC--FC relationships jointly.

\item We design a hypergraph-specific DCA fusion mechanism for
bidirectional SC--FC information exchange. Cross-modal attention over
hypergraph representations enables each modality to incorporate
complementary information while preserving modality-specific
higher-order relationships.

\item Using established topological fidelity measures — degree/edge
distributional distance, community-structure agreement (NMI, ARI),
modularity difference, and the SC--FC coupling gap — we show that
per-subject Pearson correlation does not reliably predict
topological fidelity: MHG-FM achieves the most consistent
preservation of network topology among all compared methods, even
when a baseline attains comparable or higher correlation.

\end{enumerate}

\begin{figure*}[!t]
\centering
\includegraphics[
    width=\textwidth,
    height=9cm,
    trim=0 4 0 6,
    clip
]{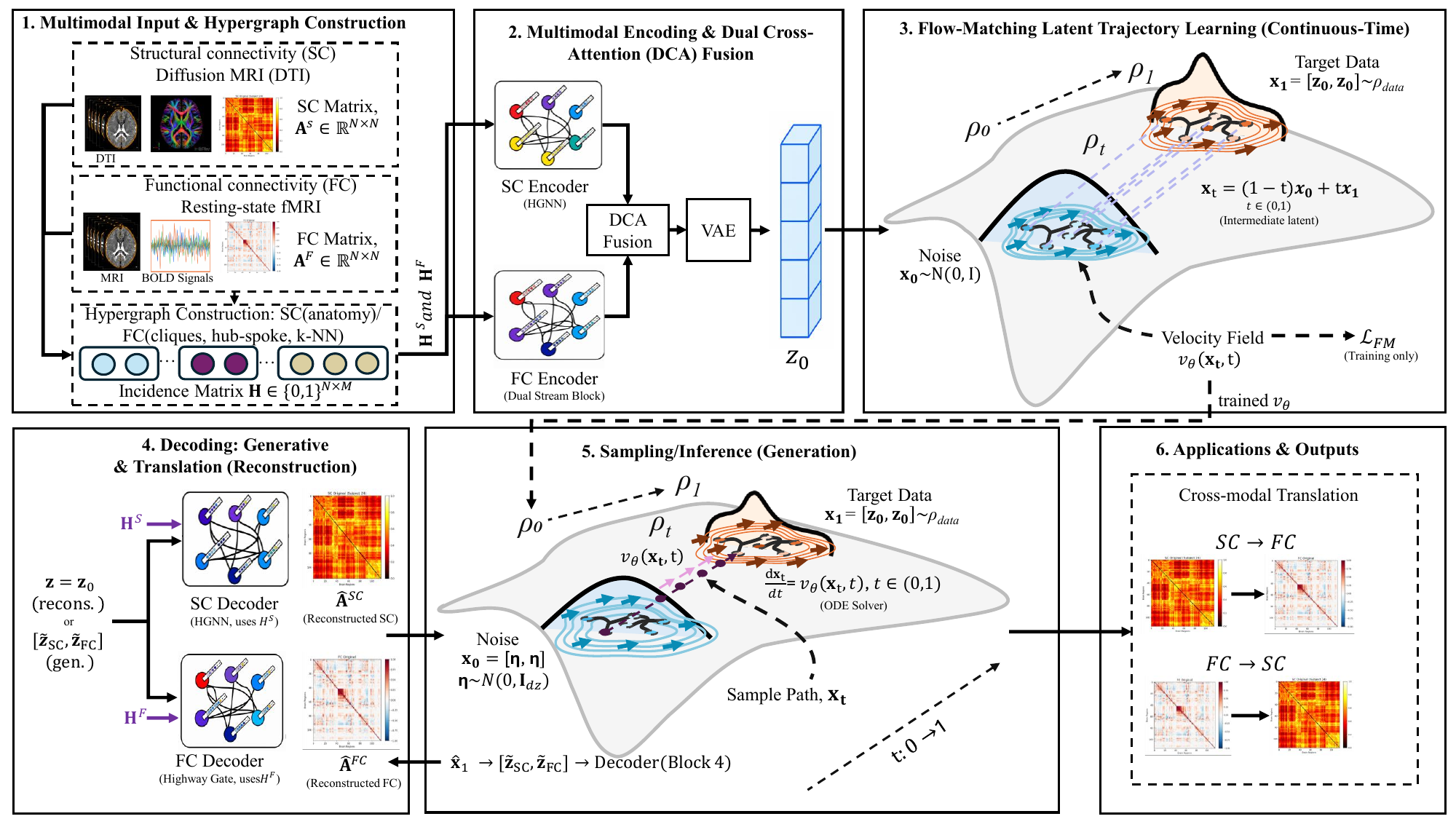}
\caption{Overview of the proposed MHG-FM framework for joint
structural-functional brain connectivity generation and translation.
Panels: (1)~Multimodal input and hypergraph construction;
(2)~Multimodal encoding and DCA fusion;
(3)~Flow-matching latent trajectory learning;
(4)~Decoding; (5)~Sampling and inference;
(6)~Applications and outputs.}
\label{fig:overall_framework}
\vspace{-2mm}
\end{figure*}

\section{METHODS}

Fig.~\ref{fig:overall_framework} shows an overview of the proposed framework.
The proposed MHG-FM represents SC and FC as separate hypergraphs, in
which each hyperedge groups multiple regions rather than a single
pair, enabling higher-order dependency modelling beyond pairwise
graphs (Section~\ref{sec:hypergraph}). Their representations are
fused through bidirectional cross-modal attention (DCA), chosen
because it directly models the well-established bidirectional
constraint between anatomical and functional connectivity rather
than assuming a fixed fusion rule (Section~\ref{sec:dca}). The joint
generative distribution is then modelled via flow matching in a
shared latent space, selected for its continuous-time formulation
that avoids the long iterative denoising chains required by
diffusion-based alternatives while retaining comparable generation
fidelity (Section~\ref{sec:fm}). Together, these design choices
enable within-modality fidelity, structure--function coupling, and
higher-order dependency modelling.

\subsection{Problem Formulation}
\label{sec:problem}

Let $\mathbf{A}^{S}\in\mathbb{R}^{N\times N}$ and
$\mathbf{A}^{F}\in\mathbb{R}^{N\times N}$ denote the weighted structural
connectivity (SC) and functional connectivity (FC) adjacency matrices,
respectively, defined over a common set of $N$ brain regions of interest
(ROIs) $\mathcal{V}$. The matrices are derived from diffusion tensor
imaging (DTI) tractography and resting-state functional MRI (rs-fMRI),
where $A^{S}_{ij}$ is the structural connection strength between ROIs
$i$ and $j$, and $A^{F}_{ij}$ is the corresponding functional
connectivity, estimated from the Pearson correlation of their BOLD
time-series. For each subject, the observed connectome is represented by
the paired adjacency matrices $(\mathbf{A}^{S},\mathbf{A}^{F})$, constituting the training data and the reconstruction targets for model training.

To capture higher-order interactions beyond pairwise connectivity, the
SC and FC networks are further represented as hypergraphs
$\mathcal{H}^{S}=(\mathcal{V},\mathcal{E}^{S},\mathbf{W}^{S})$ and
$\mathcal{H}^{F}=(\mathcal{V},\mathcal{E}^{F},\mathbf{W}^{F})$,
respectively~\cite{feng2019hgnn}, with binary incidence matrices
$\mathbf{H}^{S}\in\{0,1\}^{N\times M_{S}}$ and
$\mathbf{H}^{F}\in\{0,1\}^{N\times M_{F}}$, where $M_{S}$ and $M_{F}$
are the numbers of structural and functional hyperedges (each grouping
anatomically or functionally related ROIs, as defined in Section~\ref{sec:hypergraph}).
Following~\cite{feng2019hgnn}, all hyperedges are assigned equal
importance, i.e., $\mathbf{W}^{S}=\mathbf{I}$ and $\mathbf{W}^{F}=\mathbf{I}$.
The incidence matrices are deterministically constructed from the
observed adjacency matrices and remain fixed throughout training,
serving as structural priors for hypergraph message passing rather than
reconstruction targets.
Given the pair of connectomes $(\mathbf{A}^{S},\mathbf{A}^{F})$ alongside
their induced hypergraph representations $(\mathbf{H}^{S},\mathbf{H}^{F})$,
the objective is to learn a parameterised generative model $f_{\theta}$
that captures the joint distribution $p(\mathbf{A}^{S},\mathbf{A}^{F})$,
jointly synthesizing paired structural and functional connectomes while
preserving structure--function coupling and global topological
properties, and supporting cross-modal translation between SC and FC.

\subsection{Multimodal Hypergraph Construction}
\label{sec:hypergraph}

\textbf{SC hyperedges} are constructed by thresholding the population-mean
structural connectivity matrix $\overline{A}^S$, derived from diffusion
MRI tractography: for each region $i$, $e_i = \{i\} \cup \{j :
\overline{A}^S_{ij} > \tau_s\}$, $\tau_s = 0.01$. Since
$\overline{A}^S$ reflects physically measured white-matter connections
rather than a statistical similarity, this thresholding grounds
hyperedge membership directly in anatomical connectivity, unlike the
FC hyperedges below, which are built from statistical motifs.

\textbf{FC hyperedges} are pooled from four motif groups computed on the
population-mean FC matrix $\overline{A}^F$: (1) triangle cliques,
where region $i$ forms a hyperedge with all regions $j$ satisfying
$|\overline{A}^F_{ij}| > r_{tri} = 0.7$; (2) hub-and-spoke, where each
top 80th-percentile-degree hub connects to its $k_{star}=8$ strongest
correlates \cite{vandenheuvel2011richclub,vandenheuvel2013hubs,power2013hubs}; and (3--4) $k$-nearest-neighbour groups at $k=5$ and
$k=10$. All candidate hyperedges from these four groups, together
with the SC hyperedges above, are pooled into a single candidate
set, from which any hyperedge whose region membership exactly
duplicates one already retained is removed, yielding $\mathbf{H}^{joint}$.

A population-level joint incidence matrix
$[{\mathbf{H}^{S}},{\mathbf{H}^{F}}]$ is obtained by majority voting across training subjects.
Hyperedges appearing in at least 30\% of subjects are retained,
balancing robustness against inter-subject variability. 
The adjacency matrices $\mathbf{A}^{S}$ and $\mathbf{A}^{F}$ and the incidence matrices $\mathbf{H}^{S}$ and $\mathbf{H}^{F}$ serve different roles. SC and FC are, by
definition, pairwise measures between ROIs, so the quantities
reconstructed by the decoders and evaluated throughout
(Sections~\ref{sec:decoding}--\ref{sec:sampling}) are the adjacency
matrices $\mathbf{A}^{S}$ and $\mathbf{A}^{F}$. $\mathbf{H}^{S}$ and
$\mathbf{H}^{F}$ are not a second generative stage; they are computed
once from $\mathbf{A}^{S}$ and $\mathbf{A}^{F}$, respectively, and used as a fixed structural
scaffold within every HGNN layer of the encoder and decoder, letting the
model aggregate information over groups of related regions rather than
pairwise neighbours only (analogous to the fixed adjacency structure in a
conventional GNN) for shaping information propagation without changing the
$N\times N$ nature of the output.

\subsection{Multimodal Encoding \& Dual Cross-Attention Fusion}
\label{sec:dca}
\textbf{SC encoder} $f_{SC}$: SC matrix rows pass through two stacked
HGNN layers with residual connections and batch normalisation, then
mean-pooled to $\mathbf{z}_{SC}\in\mathbb{R}^{d_s}$. Two layers were
used to balance representational capacity against over-smoothing,
since deeper hypergraph message passing can cause node
representations to converge as more hyperedge neighbourhoods are
aggregated; residual connections preserve the original per-region
features alongside the aggregated neighbourhood information, and
batch normalisation stabilises training across the two hypergraph
propagation steps.

\textbf{FC encoder} $f_{FC}$: a dual-stream architecture with two
DualStream blocks (parallel HGNN layers with intra-block cross-attention
between node and hyperedge streams), mean-pooled to
$\mathbf{z}_{FC}\in\mathbb{R}^{d_f}$.

\textbf{DCA fusion}: The proposed DCA module
performs bidirectional information exchange between structural and
functional representations. 

It is motivated by the well-established structure-function coupling of brain connectivity, using anatomical connectivity to
constrain functional interactions and vice versa~\cite{honey2009predicting,suarez2020linking}. 
Specifically, the
cross-modal attention output from SC to FC, $\mathrm{Attn}_{SC\to FC}$,
is computed as:
\begin{equation}
  \mathrm{Attn}_{SC\to FC}=\mathrm{Softmax}\!\left(
  \frac{Q_{SC}K_{FC}^\top}{\sqrt{d}}\right)V_{FC},
\end{equation}
where $Q_{SC}$, $K_{FC}$, and $V_{FC}$ denote the query, key, and
value matrices obtained through linear projections of the SC and FC
modality embeddings, and $d$ denotes the dimensionality of the
key vectors used for attention scaling. $\mathrm{Attn}_{FC\to
SC}$ is defined analogously, with the roles of SC and FC reversed.

Following residual connections and layer normalisation, the fused
multimodal embedding is obtained by projecting the concatenated
representations through a multilayer perceptron (MLP), as follows:
\begin{equation}
  \mathbf{h}=\mathrm{MLP}([\tilde{\mathbf{z}}_{SC}, \tilde{\mathbf{z}}_{FC}, \mathbf{z}_{SC}, \mathbf{z}_{FC}])
  \in\mathbb{R}^{d_h},
\end{equation}
where $d_h=256$; the four-way concatenation preserves both attended
cross-modal and original unimodal representations.
% \subsection{Variational Latent Representation}
% \label{sec:vae}
The fused embedding $\mathbf{h}$ is subsequently projected into a shared latent
space: $q(\mathbf{z_0}|\mathbf{h}) = \mathcal{N}(\boldsymbol{\mu},
\mathrm{diag}(\boldsymbol{\sigma}^2))$, $d_z = 128$. Using the
reparameterisation trick, $\mathbf{z_0} = \boldsymbol{\mu} +
\boldsymbol{\sigma} \odot \boldsymbol{\epsilon}$, $\boldsymbol{\epsilon}
\sim \mathcal{N}(0, I)$. The Kullback--Leibler (KL) regularisation is annealed from $0$ to
$\beta_{\max} = 5\times10^{-5}$ over $9{,}000$ steps.

\subsection{Joint Multimodal Latent Flow Matching}
\label{sec:fm}
FM is a continuous-time generative modeling framework that learns a velocity field to transport samples from a simple source distribution $p_0$ to a target data distribution $p_1$. Specifically, given that $\mathbf{x}_0 \sim p_0$ denotes a sample drawn from an isotropic Gaussian (source) distribution, and let $\mathbf{x}_1=[\mathbf{z}_0, \mathbf{z}_0] \sim p_1$ denote a sample from the joint multimodal latent (target) distribution. The evolution between these distributions is defined as a time-dependent flow map $\psi_t:[0,1] \times \mathbb{R}^d \rightarrow \mathbb{R}^d$ that continuously transports samples from $p_0$ toward $p_1$, governed by the following ordinary differential equation (ODE), described as:
\begin{align} \label{eq:flow_ode}
\begin{split}
% d\Phi_t(x_0) = v_t\big(\Phi_t(x_0);\theta\big)dt, \quad x_t = \Phi_t(x_0), 
\frac{d\mathbf{x}_t}{dt} =  v_t\big(\mathbf{x}_t;\theta\big), \quad \mathbf{x}_t = \psi_t(\mathbf{x}_0),
\quad t\in[0,1],
\end{split}
\end{align}
where $\mathbf{x}_t$ denotes the intermediate state at time $t$, and $v_t(\mathbf{x}_t;\theta)$ denotes the instantaneous velocity field parameterized by $\theta$. This formulation induces a time-dependent probability path $p_{t}$. Formally, the probability path is obtained by pushing the source distribution through the flow map:
\begin{align} \label{eq:flow_matching}
\begin{split}
p_t(\mathbf{x})=[\psi_t]_*p_0(\mathbf{x})=p_0[\psi_t^{-1}(\mathbf{x})] \left|\det\left(\frac{\partial \psi_t^{-1}(\mathbf{x})}{\partial \mathbf{x}}\right)\right|,
\end{split}
\end{align}
where $[\cdot]_*$ denotes the push-forward operator and the determinant $\det$ accounts for the local change in volume induced by the flow transformation. To ensure conservation of probability mass throughout this evolution, the probability path \(p_t\) and its associated target velocity field \(u_t\) must satisfy the continuity equation:
\begin{align} \label{eq:continuity_equation}
\begin{split}
\frac{\partial p_t(\mathbf{x})}{\partial t} + \nabla_{\mathbf{x}}  \cdot \left(p_t(\mathbf{x})u_t(\mathbf{x})\right)=0 \, ,
\end{split}
\end{align}

However, directly evaluating the marginal probability path $p_t$ and its corresponding target velocity field $u_t$ is generally intractable, making direct optimization of the original flow matching objective $\mathcal{L}_{FM}
(\theta)=\mathbb{E}_{t,p_t(\mathbf{x}_t)} \big\| v_t(\mathbf{x}_t;\theta) - u_t(\mathbf{x}_t) \big\|^2_2$ impractical.
To obtain a tractable training objective, we first adopt a rectified linear interpolation path that defines the transports as $\mathbf{x}_t = (1-t) \, \mathbf{x}_0 + t \, \mathbf{x}_1$ for time step $t\in[0,1]$. The corresponding conditional target velocity field  is constructed as $u_t(\mathbf{x}_t;\mathbf{x}_1)=\frac{d\mathbf{x}_t}{dt}=\mathbf{x}_1-\mathbf{x}_0$ by differentiating the interpolation path with respect to time.
Consequently, the tractable conditional flow matching (CFM) objective can be expressed as:

\begin{align} \label{eq:conditional_flow_matching}
\begin{split}
\mathcal{L}_{CFM}
(\theta)=\mathbb{E}_{t,\mathbf{x}_0,\mathbf{x}_1} \big\| v_t(\mathbf{x}_t;\theta) - u_t(\mathbf{x}_t;\mathbf{x}_1) \big\|^2_2 .
\end{split}
\end{align}

In contrast to diffusion-based connectome generation frameworks,
which typically require hundreds to thousands of sampling steps through denoising iterations, our proposed CFM framework parameterizes the transport dynamics through a continuous-time
velocity field. This can reduce sampling complexity, while maintaining generation fidelity.

\subsection{Modality-Specific Hypergraph Decoding}
\label{sec:decoding}
The joint latent representation is decoded separately by  SC and FC decoders. Each decoder is described in detail below:

\noindent \textbf{SC decoder} $g_{SC}$: expands a $d_z$-dimensional latent vector $\mathbf{z}$ (either the reconstruction latent
$\mathbf{z}_0$ or the corresponding half of
the generative latent) into $N\times 256$ node
features via two HGNN decoder layers, followed by a symmetric pairwise
MLP head with sigmoid$\times$3 activation constraining outputs to
$(0,1)$. Given node features $\mathbf{n}_i$ and $\mathbf{n}_j$, the
reconstructed SC edge weight is
\[
\hat{\mathbf{A}}^{SC}_{ij} = \sigma\left(\mathrm{MLP}([\mathbf{n}_i, \mathbf{n}_j])\right).
\]
A symmetry constraint is enforced by averaging $\hat {\mathbf{A}}^{SC}_{ij}$ and
$\hat {\mathbf{A}}^{SC}_{ji}$.

%\noindent 
\textbf{FC decoder} $g_{FC}$: a highway gate blends local HGNN pairwise
prediction $\hat{\mathbf{A}}^{FC}_{local}$ with a global MLP prediction
$\hat{\mathbf{A}}^{FC}_{global}=\tanh(W_2\,\mathrm{ReLU}(W_1\mathbf{z}))$:
\begin{equation}
  \hat{\mathbf{A}}^{FC}=g\cdot\hat{\mathbf{A}}^{FC}_{global}
  +(1-g)\cdot\hat{\mathbf{A}}^{FC}_{local},
\end{equation}
with learned gate $g=\sigma(W_g\mathbf{z})\in(0,1)$, clamped to
$[-1,1]$.

\subsection{Training Objective}
\label{sec:training}
The final training objective consists of five components:
\begin{align}
\begin{split}
\mathcal{L}\;=&\lambda_{SC}\mathcal{L}_{SC}+\lambda_{FC}\mathcal{L}_{FC} +\lambda_{CFM}\mathcal{L}_{CFM}\\
&+\lambda_{KL}\mathcal{L}_{KL}+\lambda_{sp}\mathcal{L}_{sp},
\end{split}
\end{align}
Here, $\mathcal{L}_{SC}$ combines topology-weighted MSE
($w_{ij}=1+19\cdot\mathbf{1}[\mathbf{A}^{SC}_{ij}>0.01]$) with a Pearson
correlation auxiliary term ($\lambda_r=3.0$).
$\mathcal{L}_{FC}$ combines MSE with a Pearson auxiliary ($\lambda_r=0.5$).
$\mathcal{L}_{sp}=\frac{1}{N^2}\sum_{ij}\hat{\mathbf{A}}^{SC}_{ij}$ promotes
SC sparsity consistent with white-matter tract biology.
Loss weights ($\lambda_{SC}=7$, $\lambda_{FC}=16$, $\lambda_{FM}=0.02$,
$\lambda_{KL}=\beta_{max}=5\times10^{-5}$ with annealing,
$\lambda_{sp}=0.05$) were selected on the validation set and fixed
across all experiments, with larger weights on the modality-specific
reconstruction terms and smaller weights on flow-matching, KL, and
sparsity regularisation to stabilise optimisation without dominating
reconstruction quality.

\subsection{Multimodal Flow Matching Sampling and Generation}
\label{sec:sampling}
At inference, synthetic connectomes are generated by solving the ODE
\begin{equation}
\frac{d\mathbf{x}_t}{dt} = v_\theta(\mathbf{x}_t, t), \qquad t: 0 \to 1,
\label{eq:ode}
\end{equation}
with $\mathbf{x}_0$ initialized from a non-informative/standard Gaussian. Equivalently, integrating the velocity field gives
\[
\mathbf{x}_1 = \mathbf{x}_0 + \int_0^1 v_\theta(\mathbf{x}_t, t)\, dt.
\]
We adopt the adaptive Dormand--Prince
(\texttt{dopri5}) Runge--Kutta solver~\cite{dormand1980} with
atol=$10^{-5}$ and rtol=$10^{-4}$. The sampling is initialized from Gaussian noise
$\mathbf{x}_0=[\boldsymbol{\eta}, \boldsymbol{\eta}]$,
where $\boldsymbol{\eta}\sim\mathcal{N}(0,I_{d_z})$. The learned
velocity field is integrated from $t=0$ to $t=1$, and the resulting
$\hat{\mathbf{x}}_1$ is split into two $d_z$-dimensional halves,
$\hat{\mathbf{z}}_{SC}$ and $\hat{\mathbf{z}}_{FC}$, for
modality-specific decoding by $g_{SC}$ and $g_{FC}$. Although both halves originate from
identical noise and share the same target construction, the velocity
field's channel-wise structure allows them to diverge along the trajectory.
This divergence is an emergent property of the learned dynamics rather
than an explicit training objective.

Beyond de novo generation, the shared latent space enables
cross-modal translation without ODE-based sampling. For
FC$\to$SC translation, the unavailable SC input is zero-filled
($\mathbf{A}^S=\mathbf{0}$), while $f_{FC}$ encodes the observed FC
matrix. The two representations are fused by DCA to obtain $\mathbf{z}_0$, from which only $g_{SC}$ reconstructs the missing SC.
SC$\to$FC translation is performed analogously by zero-filling
$\mathbf{A}^F$. The shared latent representation enables
translation through a single deterministic encoder--decoder
pass, whereas de novo generation requires numerical integration
of $v_\theta$ over $t\in[0,1]$. The incidence matrices
$\mathbf{H}^S$ and $\mathbf{H}^F$ remain fixed during translation. 

To align training with this zero-filled input
distribution, the model is fine-tuned with modality dropout
($p_{\emptyset}=0.25$): each training batch presents either both
modalities (probability $1-2p_{\emptyset}$), SC zero-filled
(probability $p_{\emptyset}$), or FC zero-filled (probability
$p_{\emptyset}$), so translation at inference operates on inputs
the model has been directly trained to handle rather than an
out-of-distribution input.

\section{EXPERIMENTAL RESULTS}
\label{sec:experiments}

\subsection{Experimental Setup and Implementation Details}
\noindent\textbf{Dataset.} We use the HCP-YA 3T
dataset~\cite{vanessen2013hcp} for experimental evaluation. Informed consent
was obtained by the WU-Minn HCP Consortium at the time of data collection;
this study uses de-identified, publicly available data and did not involve
direct recruitment of participants. Resting-state fMRI was
acquired on a customised Siemens 3T Connectome Skyra scanner
(multiband EPI, factor 8; TR/TE$=720/33.1~\mathrm{ms}$; 2.0~mm isotropic
voxels), while diffusion MRI was acquired using a multi-shell protocol
($b=1000,2000,3000~\mathrm{s/mm}^2$, 90 directions per shell, 1.25~mm
isotropic voxels). Both modalities were preprocessed using the HCP Minimal
Preprocessing Pipeline~\cite{glasser2013minimal}. SC matrices were derived from
dMRI tractography and min-max normalised to $[0,1]$, whereas FC matrices were
computed as Pearson correlation matrices from rs-fMRI, followed by Fisher
$z$-transformation and clipping to $[-1,1]$. Both modalities were parcellated
using the AAL-116 atlas~\cite{tzourio2002} and split into 562/70/70
subjects for training, validation, and testing, respectively.

\begin{table*}[!t]
\centering
\caption{Connectivity-level reconstruction quality on HCP-YA 3T (paired encoder input; see Table~II for de novo
generation quality and Table~IV for sampling efficiency). Best result
in each column (per block) is bolded, second-best is underlined. $\uparrow$ higher is
better; $\downarrow$ lower is better.}
\label{tab:connectivity_matrix_3T}
\footnotesize
\setlength{\tabcolsep}{2pt}
\begin{tabular}{clccccc}
\toprule
\textbf{Connectivity} &
\textbf{Method} &
$r\uparrow$ &
MAE$\downarrow$ &
$R^2\uparrow$ &
Frob.$\downarrow$ &
$W_1\downarrow$\\
\midrule
\multirow{7}{*}{SC}
& WGAN-SC & $0.9358{\pm}0.0187$ & $0.0480{\pm}0.0078$ & $0.8356{\pm}0.0607$ & $7.0753{\pm}1.1537$ & $0.0258{\pm}0.0135$\\
& VAE-SC
& \underline{$0.9536{\pm}0.0200$}
& $0.0894{\pm}0.0286$
& $0.5004{\pm}0.2892$
& $11.3880{\pm}3.2387$
& $0.0867{\pm}0.0315$ \\
& topoGAN-SC & $0.8504{\pm}0.0141$ & $0.1144{\pm}0.0258$ & $0.0938{\pm}0.2397$ & $16.0885{\pm}2.6417$ & $0.1141{\pm}0.0261$\\
& MGCN-GAN-SC & $0.8088{\pm}0.0222$ & $0.1227{\pm}0.0175$ & $0.0049{\pm}0.3048$ & $17.4149{\pm}1.9215$ & $0.1116{\pm}0.0225$\\
& HYGENE-SC & $0.2800{\pm}0.0286$ & $0.4410{\pm}0.0249$ & $-10.6193{\pm}0.9216$ & $56.2986{\pm}2.7309$ & $0.4240{\pm}0.0319$\\
& MHG-DiT-SC (Ours)
& $\bm{0.9594{\pm}0.0164}$
& $\bm{0.0361{\pm}0.0068}$
& $\bm{0.8984{\pm}0.0394}$
& $\bm{5.3415{\pm}1.0790}$
& $\bm{0.0193{\pm}0.0104}$ \\
& MHG-FM-SC (Ours)
& $0.9526{\pm}0.0207$
& \underline{$0.0409{\pm}0.0115$}
& \underline{$0.8673{\pm}0.0713$}
& \underline{$6.0233{\pm}1.6387$}
& \underline{$0.0253{\pm}0.0149$} \\
% & MHG-DiT-SC (Ours) & $\bm{0.9594{\pm}0.0164}$ & $\bm{0.0361{\pm}0.0068}$ & $\bm{0.8984{\pm}0.0394}$ & $\bm{5.3415{\pm}1.0790}$ & $\bm{0.0193{\pm}0.0104}$\\
% & MHG-FM-SC (Ours) & \underline{$0.9566{\pm}0.0179$} & \underline{$0.0388{\pm}0.0083$} & \underline{$0.8831{\pm}0.0495$} & \underline{$5.7020{\pm}1.2208$} & \underline{$0.0233{\pm}0.0123$}\\
\midrule
\multirow{7}{*}{FC}
& WGAN-FC & $0.3228{\pm}0.0747$ & $0.1656{\pm}0.0274$ & $-0.1527{\pm}0.1996$ & $25.6458{\pm}3.6046$ & $0.0658{\pm}0.0259$\\
& VAE-FC & $0.5835{\pm}0.1405$ & $0.1212{\pm}0.0342$ & $0.2736{\pm}0.2056$ & $18.6203{\pm}4.9847$ & $0.0774{\pm}0.0355$\\
& topoGAN-FC & $0.3508{\pm}0.0674$ & $0.3137{\pm}0.0459$ & $-3.0378{\pm}1.1442$ & $44.2566{\pm}5.6911$ & $0.3118{\pm}0.0463$\\
& MGCN-GAN-FC & $0.1928{\pm}0.0969$ & $0.1528{\pm}0.0306$ & $-0.0360{\pm}0.0710$ & $24.4464{\pm}4.3000$ & $0.1470{\pm}0.0305$\\
& HYGENE-FC & $0.1342{\pm}0.0450$ & $0.8728{\pm}0.0418$ & $-26.8137{\pm}9.8080$ & $110.6517{\pm}4.7477$ & $0.8204{\pm}0.0552$\\
& MHG-DiT-FC (Ours)
& \underline{$0.8321{\pm}0.0388$}
& \underline{$0.0846{\pm}0.0210$}
& \underline{$0.6565{\pm}0.0801$}
& \underline{$16.7716{\pm}2.2644$}
& $\bm{0.0531{\pm}0.0232}$ \\
& MHG-FM-FC (Ours)
& $\bm{0.8359{\pm}0.0450}$
& $\bm{0.0834{\pm}0.0185}$
& $\bm{0.6625{\pm}0.0911}$
& $\bm{16.6185{\pm}2.0086}$
& \underline{$0.0541{\pm}0.0210$} \\
% & MHG-DiT-FC (Ours) & $\bm{0.8321{\pm}0.0388}$ & \underline{$0.0846{\pm}0.0210$} & \underline{$0.6565{\pm}0.0801$} & \underline{$16.7716{\pm}2.2644$} & \underline{$0.0531{\pm}0.0232$}\\
% & MHG-FM-FC (Ours) & \underline{$0.8318{\pm}0.0476$} & $\bm{0.0828{\pm}0.0200}$ & $\bm{0.6664{\pm}0.0880}$ & $\bm{16.5983{\pm}2.1795}$ & $\bm{0.0476{\pm}0.0210}$\\
\midrule
\multirow{14.5}{*}{FC+SC}
& WGAN-Joint (SC) & $0.9347{\pm}0.0220$ & $0.0594{\pm}0.0213$ & $0.7489{\pm}0.1817$ & $8.4997{\pm}2.7814$ & $0.0440{\pm}0.0251$\\
& VAE-Joint (SC) & $0.9555{\pm}0.0209$ & $0.1092{\pm}0.0354$ & $0.2883{\pm}0.4336$ & $13.5436{\pm}4.0059$ & $0.1078{\pm}0.037$\\
& topoGAN-Joint (SC) & $0.8479{\pm}0.0145$ & $0.1167{\pm}0.0259$ & $0.0630{\pm}0.2435$ & $16.3624{\pm}2.6492$ & $0.1166{\pm}0.0260$\\
& MGCN-GAN-Joint (SC) & $0.5676{\pm}0.0321$ & $0.1020{\pm}0.0065$ & $0.2438{\pm}0.0755$ & $15.6352{\pm}1.0949$ & $0.0592{\pm}0.0144$\\
& HYGENE-Joint (SC) & $0.2851{\pm}0.0242$ & $0.4339{\pm}0.0237$ & $-10.5146{\pm}0.87155$ & $56.0394{\pm}2.4595$ & $0.4014{\pm}0.0254$\\
& MHG-DiT-Joint (SC) (Ours) & $\bm{0.9593{\pm}0.0183}$ & $\bm{0.0366{\pm}0.0064}$ & $\bm{0.8961{\pm}0.0426}$ & $\bm{5.3798{\pm}1.0292}$ & $\bm{0.0208{\pm}0.0099}$\\
& MHG-FM-Joint (SC) (Ours)
& \underline{$0.9565{\pm}0.0222$}
& \underline{$0.0383{\pm}0.0079$}
& \underline{$0.8856{\pm}0.0529$}
& \underline{$5.6250{\pm}1.2391$}
& \underline{$0.0226{\pm}0.0115$} \\
% & MHG-DiT-Joint (SC) (Ours) & $\bm{0.9593{\pm}0.0183}$ & $\bm{0.0366{\pm}0.0064}$ & $\bm{0.8961{\pm}0.0426}$ & $\bm{5.3798{\pm}1.0292}$ & $\bm{0.0208{\pm}0.0099}$\\
% & MHG-FM-Joint (SC) (Ours) & \underline{$0.9584{\pm}0.0206$} & \underline{$0.0370{\pm}0.0067$} & \underline{$0.8936{\pm}0.0462$} & \underline{$5.4441{\pm}1.1041$} & \underline{$0.0211{\pm}0.0097$}\\
\cmidrule{2-7}
& WGAN-Joint (FC) & $0.3553{\pm}0.0785$ & $0.1734{\pm}0.0471$ & $-0.3341{\pm}0.6754$ & $27.0169{\pm}6.0482$ & $0.0942{\pm}0.0502$\\
& VAE-Joint (FC) & $0.5911{\pm}0.1417$ & $0.1212{\pm}0.0334$ & $0.2765{\pm}0.2263$ & $18.5152{\pm}4.8892$ & $0.0716{\pm}0.0327$\\
& topoGAN-Joint (FC) & $0.3496{\pm}0.0700$ & $0.3223{\pm}0.0459$ & $-3.2193{\pm}0.2072$ & $45.1536{\pm}5.7020$ & $0.3215{\pm}0.0461$\\
& MGCN-GAN-Joint (FC) & $0.4484{\pm}0.1308$ & $0.1635{\pm}0.0212$ & $-0.1897{\pm}0.3928$ & $24.4222{\pm}2.9480$ & $0.1207{\pm}0.0212$\\
& HYGENE-Joint (FC) & $0.1335{\pm}0.0374$ & $0.8772{\pm}0.0370$ & $-27.1852{\pm}10.3913$ & $111.1999{\pm}4.5529$ & $0.7969{\pm}0.0434$\\
& MHG-DiT-Joint (FC) (Ours)
& \underline{$0.8136{\pm}0.0623$}
& \underline{$0.0855{\pm}0.0226$}
& \underline{$0.6402{\pm}0.1061$}
& \underline{$17.0033{\pm}2.5577$}
& \underline{$0.0487{\pm}0.0223$} \\
& MHG-FM-Joint (FC) (Ours)
& $\bm{0.8278{\pm}0.0358}$
& $\bm{0.0834{\pm}0.0199}$
& $\bm{0.6615{\pm}0.0745}$
& $\bm{16.6926{\pm}2.1193}$
& $\bm{0.0475{\pm}0.0216}$ \\

% & MHG-DiT-Joint (FC) (Ours) & \underline{$0.8136{\pm}0.0623$} & \underline{$0.0855{\pm}0.0226$} & \underline{$0.6402{\pm}0.1061$} & \underline{$17.0033{\pm}2.5577$} & $\bm{0.0487{\pm}0.0223}$\\
% & MHG-FM-Joint (FC) (Ours) & $\bm{0.8185{\pm}0.0425}$ & $\bm{0.0851{\pm}0.0209}$ & $\bm{0.6404{\pm}0.0875}$ & $\bm{16.9656{\pm}2.3039}$ & \underline{$0.0497{\pm}0.0228$}\\
\bottomrule
\end{tabular}
\vspace{-2mm}
\end{table*}

% & topoGAN-SC & $0.8719{\pm}0.0107$ & $0.0780{\pm}0.0206$ & $0.4978{\pm}0.1581$ & $11.9507{\pm}2.2282$ & $0.0705{\pm}0.0263$\\
% & MGCN-GAN-SC & $0.7839{\pm}0.0442$ & $0.1227{\pm}0.0059$ & $-0.0311{\pm}0.0418$ & $17.9772{\pm}0.8867$ & $0.1227{\pm}0.0059$\\

% \rowcolor{ablationrow} MHG-BCA (Ours) & $0.9525{\pm}0.0185$ & $0.0419{\pm}0.0084$ & $0.8724{\pm}0.0499$ & $6.2443{\pm}1.2491$ & $0.0263{\pm}0.0121$ & $0.6212{\pm}0.1123$ & $0.1257{\pm}0.0743$ & $0.2272{\pm}0.5105$ & $20.9456{\pm}8.9942$ & $0.0817{\pm}0.0795$ \\
% \rowcolor{ablationrow} MHG-BCA-Score (Ours) & $0.9582{\pm}0.0165$ & \underline{$0.0370{\pm}0.0060$} & \underline{$0.8941{\pm}0.0404$} & \underline{$5.4421{\pm}0.9658$} & \underline{$0.0210{\pm}0.0095$} & $0.8153{\pm}0.0582$ & \underline{$0.0846{\pm}0.0204$} & $0.6454{\pm}0.0940$ & $16.8799{\pm}2.1935$ & $0.0512{\pm}0.0205$ \\
% \rowcolor{ablationrow} MHG-DCA (Ours)& $0.9578{\pm}0.0189$ & $0.0385{\pm}0.0082$ & $0.8843{\pm}0.0534$ & $5.6533{\pm}1.2294$ & $0.0231{\pm}0.0128$ & \underline{$0.8210{\pm}0.0505$} & $\bm{0.0828{\pm}0.0183}$$ & \underline{$0.6566{\pm}0.0814$} & $\bm{16.7168{\pm}1.9353}$ & $\bm{0.0442{\pm}0.0199}$\\

\noindent\textbf{Network and Hyperparameters.} The HGNN hidden dimensions are $[256,128,128]$ for SC and $256$ for FC, with a latent dimension of $d_z=128$ and a DCA dimension of $d_h=256$. The DCA module uses four blocks and four attention heads, with a dropout rate of $0.1$. Models are trained using AdamW~\cite{loshchilov2019} with a learning rate of $3\times10^{-4}$, weight decay of $10^{-3}$, an 8-epoch warmup followed by cosine annealing, a batch size of 16, and a maximum of 250 epochs. Early stopping is applied with a patience of 60 epochs, and all experiments use seed~42. 

The reported $\pm$ values throughout this section represent variability across the $n_{\mathrm{test}}=70$ test subjects for a single trained model, rather than variability across independently trained model instances. 

\noindent\textbf{Methods for comparison.} WGAN-SC/FC/Joint are Wasserstein GANs with graph-convolutional generators and discriminators and gradient-penalty regularisation, adapted for unimodal and joint generation. topoGAN~\cite{bessadok2021} and MGCN-GAN~\cite{zhang2020mgcngan,zhang2022mgcngan} were originally developed for conditional prediction. For unconditional generation, we replace their conditioning inputs with $\mathbf{z}\sim\mathcal{N}(0,\mathbf{I})$ while retaining their original architectures and training objectives. VAE-SC/FC/Joint use graph-convolutional encoders and decoders with a KL-regularised latent space. HYGENE~\cite{gailhard2025hygene} is a hierarchical diffusion-based hypergraph generator that progressively coarsens and refines a bipartite node--hyperedge representation. All baselines use the same train/validation/test split as the proposed method.

\noindent\textbf{Variants of proposed method.}
We evaluate the proposed framework through component and architecture ablations. MHG-A1 removes SC--FC cross-attention from the velocity backbone while retaining DCA fusion. MHG-A2 replaces AdaLN time conditioning with additive conditioning while retaining DCA and flow matching. MHG-Score replaces the flow-matching backbone with a score-based generative backbone, while MHG-GNN replaces the HGNN encoders with pairwise graph convolution. MHG-SC and MHG-FC denote the unimodal pathways. MHG-DiT uses a diffusion-transformer backbone~\cite{peebles2023dit}, whereas MHG-FM is the full proposed architecture with DCA fusion and a flow-matching backbone~\cite{lipman2023}.

\noindent\textbf{Evaluation Metrics.} Matrix-level metrics include Pearson $r$, MAE, $R^2$, Frobenius norm, and Wasserstein-1 distance ($W_1$)~\cite{suarez2020linking,you2018graphrnn}. Topology-level metrics include Degree/Edge $W_1$,  normalized mutual information (NMI) and adjusted
Rand index (ARI) for community-structure agreement, and the SC--FC
matching-index gap $|\Delta\mathrm{MI}|$~\cite{rubinov2010,vignac2023digress,taylor2026,suarez2020linking}. The matching index is estimated using 20-bin histograms of upper-triangle weights. Since unconditional generation does not preserve subject-wise correspondence,
NMI and ARI are averaged over random permutations of generated samples
relative to the real test samples. For $K$ random permutations
$\{\pi_k\}_{k=1}^{K}$, the permutation-averaged NMI is
\begin{equation}
\begin{aligned}
\overline{\mathrm{NMI}}
&=
\frac{1}{K N_{\mathrm{test}}}
\sum_{k=1}^{K}
\sum_{i=1}^{N_{\mathrm{test}}}
\mathrm{NMI}
\left(
C_i^{\mathrm{real}},
C_{\pi_k(i)}^{\mathrm{gen}}
\right).
\end{aligned}
\end{equation}
with ARI computed analogously. The same procedure is applied to all
compared methods.

\subsection{Generation Quality}
\label{sec:matrix-level}

\begin{table*}[!t]
\centering
\caption{NMI and ARI are computed by applying data-driven community
detection independently to each generated sample and each real
test-subject sample. Because generation is unconditional
($\mathbf{z}\sim\mathcal{N}(0,\mathbf{I})$), generated samples have no
subject-wise correspondence with the real test subjects. To obtain a
population-level estimate independent of arbitrary sample ordering,
NMI and ARI are therefore averaged over repeated random permutations
of the generated samples relative to the real samples, using the same
procedure for all compared methods. $|\Delta\mathrm{MI}|$ is defined
only for jointly generated SC--FC pairs and is reported once per
method, spanning both its SC and FC rows. Best result in each column
(per block) is bolded; second-best is underlined.}
\label{tab:network_topology_3T}
\footnotesize
\setlength{\tabcolsep}{8pt}
\begin{tabular}{clcccccc}
\toprule
\textbf{Connectivity} &
\textbf{Method} &
Deg-$W_1\downarrow$ &
Edge-$W_1\downarrow$ &
\textbf{NMI}$\uparrow$ &
\textbf{ARI}$\uparrow$ &
$|\Delta\mathrm{MI}|\downarrow$ \\
\midrule
\multirow{7}{*}{SC}
& WGAN-SC
& \underline{$0.5411$}
& $\bm{0.0117}$
& $0.4029$
& \underline{$0.3921$}
& -- \\
& VAE-SC
& $0.6815$
& $0.0915$
& $\bm{0.4109}$
& $\bm{0.3942}$
& -- \\
& topoGAN-SC
& $1.2719$
& $0.1191$
& \underline{$0.4085$}
& $0.3540$
& -- \\
& MGCN-GAN-SC
& $2.2099$
& $0.1099$
& $0.3554$
& $0.3525$
& -- \\
& HYGENE-SC
& $5.5510$
& $0.4233$
& $0.2791$
& $0.1679$
& -- \\
& MHG-DiT-SC (Ours)
& $10.0485$
& $0.5218$
& $0.2351$
& $0.0536$
& -- \\
& MHG-FM-SC (Ours)
& $\bm{0.4980}$
& \underline{$0.0135$}
& $0.3905$
& $0.3599$
& -- \\
\midrule
\multirow{7}{*}{FC}
& WGAN-FC
& $6.9192$
& \underline{$0.0480$}
& $0.1984$
& $0.0941$
& -- \\
& VAE-FC
& \underline{$1.4640$}
& $0.0724$
& \underline{$0.3667$}
& $\bm{0.1664}$
& -- \\
& topoGAN-FC
& $5.3254$
& $0.3292$
& $0.2553$
& $0.0549$
& -- \\
& MGCN-GAN-FC
& $1.6761$
& $0.1468$
& $0.1850$
& $0.0611$
& -- \\
& HYGENE-FC
& $5.9089$
& $0.8152$
& $0.2402$
& $-0.0346$
& -- \\
& MHG-DiT-FC (Ours)
& $8.8914$
& $0.8443$
& $0.2583$
& $-0.0138$
& -- \\
& MHG-FM-FC (Ours)
& $\bm{1.2985}$
& $\bm{0.0418}$
& $\bm{0.3713}$
& \underline{$0.1321$}
& -- \\
\midrule
\multirow{14.5}{*}{FC+SC}
& WGAN-Joint (SC)
& \underline{$0.4835$}
& $\bm{0.0113}$
& $0.4063$
& \underline{$0.4001$}
& \underline{$0.0122$}
&  \\
& VAE-Joint (SC)
& $0.6473$
& $0.1075$
& $0.3967$
& $0.3546$
& $0.0476$
&  \\
& topoGAN-joint (SC)
& $1.2741$
& $0.1191$
& \underline{$0.4085$}
& $0.3540$
& $0.2006$
& \\
& MGCN-GAN-joint (SC)
& $12.8335$
& $0.0584$
& $0.2555$
& $0.0076$
& $0.1023$
& \\
& HYGENE-joint (SC)
& $5.6512$
& $0.3997$
& $0.2938$
& $0.1841$
& $0.8125$
& \\
& MHG-DiT (SC) (Ours)
& $6.4909$
& $0.1074$
& $0.1781$
& $0.0828$
& \underline{$0.0072$}
& \\
& MHG-FM (SC) (Ours)
& $\bm{0.4640}$
& \underline{$0.0122$}
& $\bm{0.4237}$
& $\bm{0.4053}$
& $0.0188$
& \\
\cmidrule{2-8}
& WGAN-Joint (FC)
& $6.5323$
& \underline{$0.0441$}
& $0.2059$
& $0.1024$
& \underline{$0.0122$}
& \\
& VAE-Joint (FC)
& \underline{$1.3648$}
& $0.0650$
& \underline{$0.3604$}
& $\bm{0.1607}$
& $0.0476$
&  \\
& topoGAN-joint (FC)
& $5.2741$
& $0.3293$
& $0.1315$
& $0.0305$
& $0.2006$
&  \\
& MGCN-GAN-joint (FC)
& $5.5424$
& $0.1114$
& $0.3600$
& $0.0813$
& $0.1023$
&  \\
& HYGENE-joint (FC)
& $6.4103$
& $0.7817$
& $0.2461$
& $-0.0301$
& $0.8125$
&  \\
& MHG-DiT (FC) (Ours)
& $6.6500$
& $0.0765$
& $0.2280$
& $0.0779$
& $\bm{0.0072}$
&  \\
& MHG-FM (FC) (Ours)
& $\bm{0.5788}$
& $\bm{0.0368}$
& $\bm{0.3622}$
& \underline{$0.1154$}
& $0.0188$
&  \\
\bottomrule
\end{tabular}
\vspace{-2mm}
\end{table*}

Table~\ref{tab:connectivity_matrix_3T} and Table~\ref{tab:network_topology_3T}
report matrix- and topology-level generation quality of different methods on HCP-YA 3T dataset.

\noindent\textbf{SC.} MHG-DiT achieves the highest SC accuracy across the evaluated settings,
with MHG-FM showing comparable performance. 

Among the external baselines, WGAN-SC and
VAE-SC achieve Pearson $r$ values comparable to MHG-FM-SC but substantially
lower $R^2$, indicating that they capture the overall connectivity pattern
but reproduce absolute connectivity values less accurately. topoGAN-SC,
MGCN-GAN-SC and HYGENE-SC perform worse under the unconditional setting, which
may reflect their original conditional design.

\noindent\textbf{FC.} MHG-FM achieves the highest FC accuracy among the
multimodal methods in both settings, with MHG-DiT showing comparable
performance. MHG-FM has a higher $W_1$ in the unimodal setting, whereas
MHG-DiT performs better in the joint setting. No external baseline
exceeds an $R^2$ of $0.28$, WGAN-FC,
topoGAN-FC, MGCN-GAN-FC, and HYGENE-FC produce negative R2 values. The larger performance gap relative to SC
is consistent with the greater inter-subject variability of FC, making
population-level unconditional generation more challenging
~\cite{finn2015fingerprint}. VAE-FC performs comparatively better among
the external baselines.

\noindent\textbf{Topology.} Deg-$W_1$ and Edge-$W_1$ measure
distributional differences in degree and edge-weight structure without
requiring subject-wise correspondence. NMI and ARI therefore reflect
population-level community-structure agreement rather than
subject-specific reconstruction. MHG-FM achieves the strongest overall
topology preservation, with the lowest Deg-$W_1$ for both SC and FC and
the best Edge-$W_1$ and NMI for FC. WGAN and VAE remain competitive on
selected SC community-structure measures, whereas topoGAN shows larger
degree and edge-distribution discrepancies and MGCN-GAN and HYGENE show
weaker community-structure agreement. MHG-DiT exhibits substantially
weaker topology despite its strong matrix-level SC fidelity, indicating
that accurate connectivity values do not necessarily preserve
population-level network organization. For SC--FC coupling, MHG-DiT
achieves the smallest $|\Delta\mathrm{MI}|$, while MHG-FM outperforms
all external joint baselines except WGAN-Joint. Overall, MHG-FM provides
the most consistent topology preservation across both modalities while
maintaining competitive matrix fidelity and coupling.

\begin{figure*}[!tbp]
\centering
\includegraphics[width=0.86\textwidth]{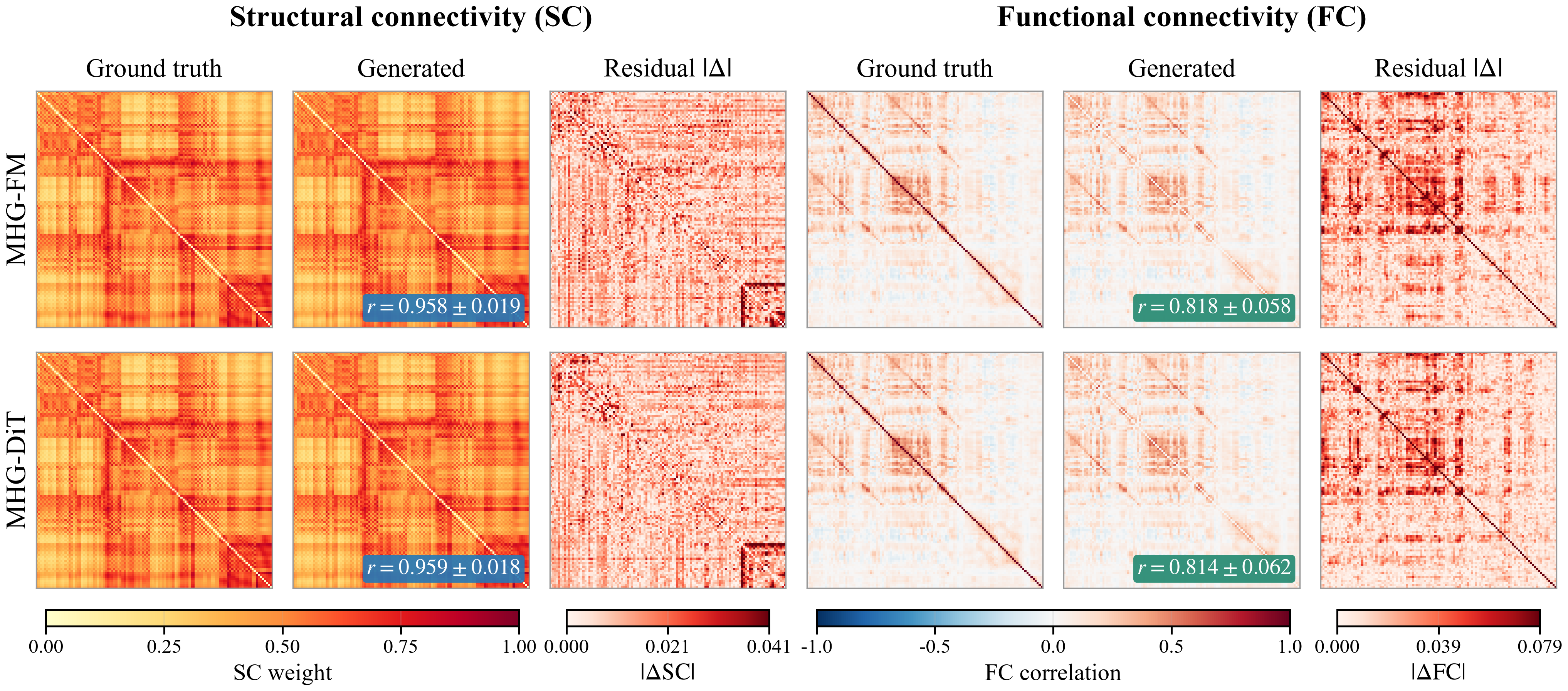}
\caption{Ground truth versus generated SC and FC matrices, averaged
across all $n{=}70$ test subjects, for MHG-FM~(top) and
MHG-DiT~(bottom). Columns: SC ground truth/generated/residual, FC
ground truth/generated/residual.}
\label{fig:qualitative}
\vspace{-2mm}
\end{figure*}

\noindent\textbf{Qualitative.} Fig.~\ref{fig:qualitative} shows SC and FC
matrices averaged across the $70$ test subjects. Both backbones recover
the modular SC organisation and tract-density gradients, with residual
errors concentrated around a small number of high-degree edges. They also
capture the block-diagonal FC structure and task-positive/negative
anti-correlation, without an apparent systematic spatial pattern in the
residuals.

% The matrix-level results show no uniform winner: 
At the matrix-level, MHG-DiT performs better
on selected SC metrics, whereas MHG-FM performs better on FC reconstruction
and cross-modal translation (Table~\ref{tab:connectivity_matrix_3T}).
At the topology level, however, MHG-FM provides more consistent preservation
of degree, edge-weight, and community structure than MHG-DiT. The comparatively weak performance of HYGENE may partly due to its original design as a general-purpose hypergraph generator rather than a multimodal structural-functional model.

The relatively low inter-subject variability of SC~\cite{hagmann2008core}
may allow
population-level generation to achieve high Pearson $r$ without fully
capturing subject-specific structure. Thus, SC $r$ should be interpreted
alongside $R^2$ and the topology-level metrics in Table~\ref{tab:network_topology_3T} rather than
in isolation. 
The lower FC correlation is consistent with its greater inter-subject
variability~\cite{finn2015fingerprint}, which makes population-level
generation more challenging. The modest reduction in FC correlation for MHG-FM relative
to MHG-FM-FC reflects a trade-off between unimodal accuracy and the joint
generation and SC--FC coupling targeted by the proposed framework.

More broadly, the gap between MHG-DiT's matrix-level and topology-level
performance indicates that Pearson $r$ alone is insufficient to evaluate
generative connectomes. MHG-FM provides a more favourable balance between
matrix fidelity and topology preservation while reducing sampling cost
relative to the matched diffusion backbone~\cite{lipman2023,liu2023rectifiedflow}
(Section~\ref{sec:efficiency}). WGAN-Joint remains competitive on SC
topology but performs markedly worse on FC.

% =============================================================================
% Table 3 — Ablation Study (Reconstruction Mode)
% All metrics: paired encoder mode (full_test_metrics)
% HCP-YA 3T, without augmentation, n=70 test subjects
% SC-FC coupling via |ΔMI| (Taylor et al. 2026)
% =============================================================================
% =============================================================================
% Table 3 — Ablation Study (Reconstruction Mode)
% All metrics: paired encoder mode (full_test_metrics)
% HCP-YA 3T, without augmentation, n=70 test subjects
% SC-FC coupling via |ΔMI| (Taylor et al. 2026)
% =============================================================================
% =============================================================================
% Table 3 — Ablation Study (Reconstruction Mode)
% All metrics: paired encoder mode (full_test_metrics)
% HCP-YA 3T, without augmentation, n=70 test subjects
% SC-FC coupling via |ΔMI| (Taylor et al. 2026)
% =============================================================================
% =============================================================================
% Table 3 — Ablation Study (Reconstruction Mode)
% All metrics: paired encoder mode (full_test_metrics)
% HCP-YA 3T, without augmentation, n=70 test subjects
% SC-FC coupling via |ΔMI| (Taylor et al. 2026)
% =============================================================================
\begin{table*}[!t]
\centering
\caption{
Ablation study on HCP-YA 3T ($n_{\mathrm{test}}=70$), paired-encoder
reconstruction mode. Higher is better for $r$ and $R^2$; lower is
better for MAE and $|\Delta\mathrm{MI}|$~\cite{taylor2026,suarez2020linking}.
Bold indicates the best result among the five joint (SC+FC) variants
(generative-backbone and encoder groups); the unimodal rows are shown
for reference and excluded from this ranking, since removing a
modality changes what is being measured. See notes below the table
for a brief description of each variant.
}
\label{tab:ablation}
\setlength{\tabcolsep}{1.5pt}
\renewcommand{\arraystretch}{1.15}
\footnotesize
\begin{tabular}{l ccc ccc c}
\toprule
&
\multicolumn{3}{c}{\textbf{SC}} &
\multicolumn{3}{c}{\textbf{FC}} &
\textbf{Coupling} \\
\cmidrule(lr){2-4}\cmidrule(lr){5-7}\cmidrule(lr){8-8}
\textbf{Model}
  & $r\,\uparrow$ & MAE$\,\downarrow$ & $R^2\,\uparrow$
  & $r\,\uparrow$ & MAE$\,\downarrow$ & $R^2\,\uparrow$
  & $|\Delta\mathrm{MI}|\,\downarrow$ \\
\midrule
\multicolumn{8}{c}{\textit{Unimodal variants}} \\
\midrule
MHG-FM-SC
  & $0.9526{\pm}0.0207$ & $0.0409{\pm}0.0115$ & $0.8673{\pm}0.0713$
  & \textit{n/a} & \textit{n/a} & \textit{n/a}
  & -- \\
MHG-FM-FC
  & \textit{n/a} & \textit{n/a} & \textit{n/a}
  & $0.8359{\pm}0.0450$ & $0.0834{\pm}0.0185$ & $0.6625{\pm}0.0911$
  & -- \\
\midrule
\multicolumn{8}{c}{\textit{Generative-backbone variants}} \\
\midrule
MHG-A1
  & \bm{$0.9590{\pm}0.0203$} & $0.0371{\pm}0.0073$ & $0.8928{\pm}0.0480$
  & $0.8118{\pm}0.0470$ & $0.0867{\pm}0.0207$ & $0.6286{\pm}0.0946$
  & $0.0258$ \\
MHG-A2
  & $0.9533{\pm}0.0291$ & $0.0402{\pm}0.0100$ & $0.8710{\pm}0.0806$
  & $0.8251{\pm}0.0457$ & $0.0847{\pm}0.0204$ & $0.6443{\pm}0.0963$
  & $0.0256$ \\
MHG-Score
  & $0.9584{\pm}0.0206$ & \bm{$0.0363{\pm}0.0064$} & \bm{$0.8971{\pm}0.0448$}
  & $0.8267{\pm}0.0404$ & $0.0834{\pm}0.0204$ & $0.6558{\pm}0.0828$   & $0.0268$  \\
% \bm{MHG-FM (Ours)}
MHG-FM
  & $0.9565{\pm}0.0222$ & $0.0383{\pm}0.0079$ & $0.8856{\pm}0.0529$
  & \bm{$0.8278{\pm}0.0358$} & \bm{$0.0834{\pm}0.0199$} & \bm{$0.6615{\pm}0.0745$}
  & \bm{$0.0188$} \\
\midrule
\multicolumn{8}{c}{\textit{Encoder variant}} \\
\midrule
MHG-GNN
  & $0.9510{\pm}0.0195$ & $0.0415{\pm}0.0081$ & $0.8659{\pm}0.0549$
  & $0.8181{\pm}0.0509$ & $0.0879{\pm}0.0206$ & $0.6259{\pm}0.1230$
  & $0.0316$ \\
\bottomrule
\end{tabular}
\vspace{4pt}

\parbox{\textwidth}{%
\footnotesize
\textbf{Notes:}
MHG-FM-SC/FC: SC-only/FC-only pathway (no other modality's encoder or decoder).
MHG-A1: no cross-attention between SC and FC streams in the velocity backbone.
MHG-A2: additive time conditioning in place of AdaLN.
MHG-Score: score-based backbone in place of flow matching.
MHG-FM (Ours): full DCA fusion with a flow-matching backbone (proposed).
MHG-GNN: pairwise graph convolution in place of the HGNN encoders.
$\Delta MI$: SC--FC coupling gap via mutual information on reconstructed pairs
\cite{taylor2026}; unimodal rows have no computed gap since one modality is not generated.
}
\vspace{-2mm}
\end{table*}
\subsection{Ablation Study}
\label{sec:ablation}

Table~\ref{tab:ablation} shows ablation of each component, encoder, and backbone of the proposed framework for the reconstruction mode with paired encoder inputs.
The unimodal pathways, MHG-FM-SC and MHG-FM-FC have no coupling gap by construction.
% and are included for reference.
Among the joint variants, MHG-A1 slightly improves SC accuracy over
MHG-FM but yields lower FC accuracy and weaker coupling, highlighting
the benefit of cross-attention for cross-modal information exchange. MHG-A2 performs worse on all FC metrics, indicating that AdaLN provides
more effective time conditioning than additive conditioning. MHG-Score achieves the best SC MAE and $R^2$ among the joint variants,
though MHG-A1 attains a marginally higher SC $r$; MHG-Score is also
closest to MHG-FM on FC, but has a larger coupling gap.
Despite lower SC accuracy than MHG-A1 and MHG-Score, MHG-FM is the only
joint variant that performs best simultaneously on FC accuracy and
coupling, with the lowest $|\Delta\mathrm{MI}|$ among all variants.

Replacing the HGNN encoders with standard pairwise graph convolution,
MHG-GNN underperforms MHG-FM across all metrics in Table~\ref{tab:ablation},
demonstrating the benefit of hypergraph message passing under the same
training setup.
The unimodal pathways cannot be evaluated for SC--FC coupling because
they generate only one modality, highlighting the need for joint
multimodal generation to evaluate structure--function coupling.

\subsection{Training Stability and Statistical Significance}
\label{sec:stability}

MHG-FM was evaluated with three random seeds (42, 123, and 456) using
identical hyperparameters. Seed 42 was selected for the main experiments
based on validation performance. Across seeds, SC Pearson $r$ was 0.9584, 0.9562, and 0.9573, and FC Pearson $r$ was 0.8165, 0.8063, and 0.8187,yielding means$\pm$SD of 0.9573$\pm$0.0009 and
0.8138$\pm$0.0054, respectively. These variations are substantially
smaller than the subject-level variability reported in Table~I
(SC SD$\approx$0.018--0.021; FC SD$\approx$0.044--0.050), indicating that the $\pm$ values reported
throughout the paper primarily reflect inter-subject variability rather
than training instability.

To assess whether the reconstruction differences between MHG-FM and
MHG-DiT in Table~\ref{tab:connectivity_matrix_3T} exceed per-subject
noise, we applied paired Wilcoxon signed-rank tests to the $n=70$
per-subject metrics under the paired-encoder condition used in
Tables~\ref{tab:connectivity_matrix_3T} and~\ref{tab:ablation}, with
Holm--Bonferroni correction across eleven comparisons
~\cite{holm1979simple}. The resulting aggregate values agree with
Table~\ref{tab:connectivity_matrix_3T} to within 0.0028 for SC $R^2$,
which is below the maximum cross-seed variability of 0.0054.

\begin{figure}[!t]
\centering
\includegraphics[width=0.9\columnwidth]{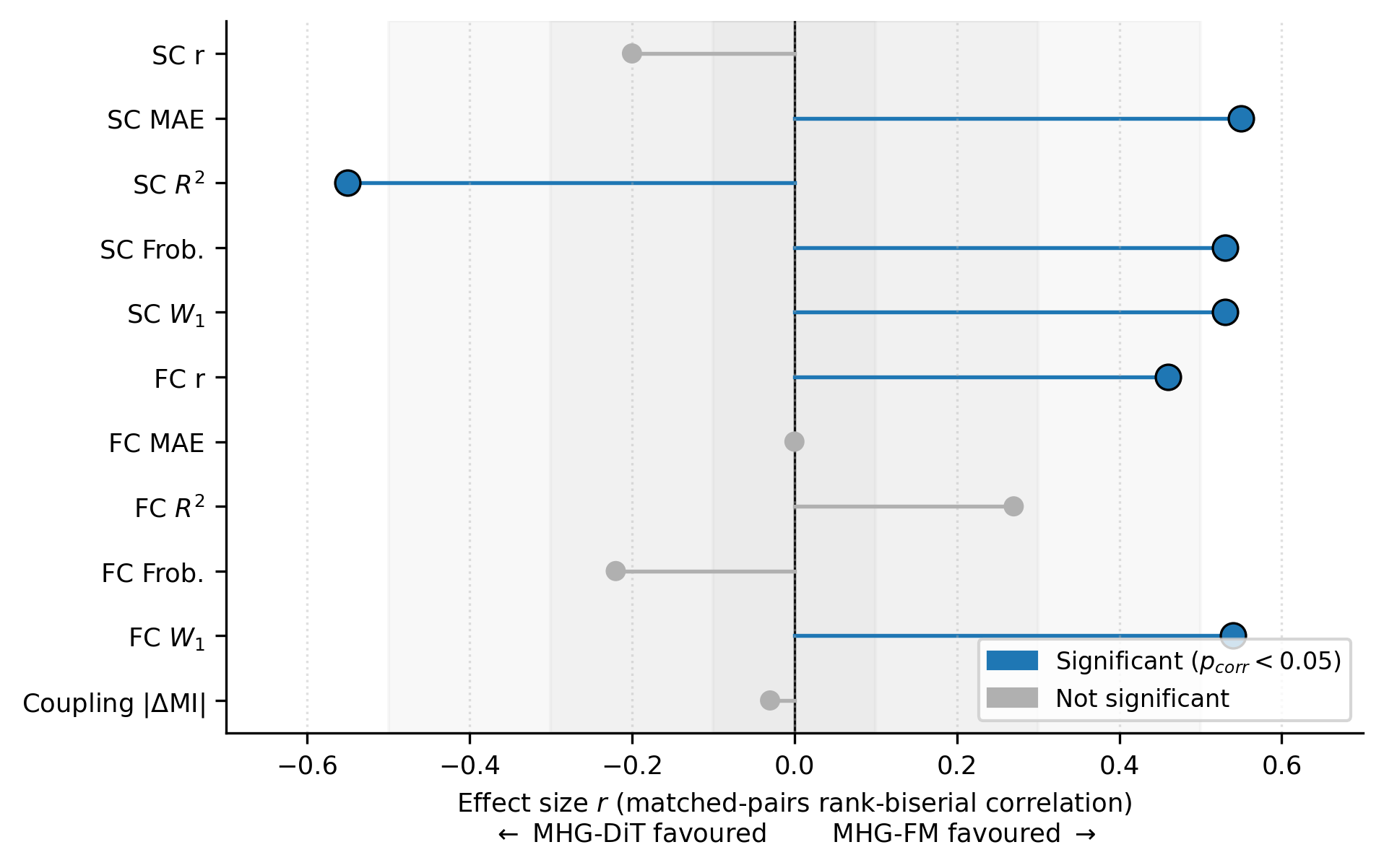}
\caption{
Effect sizes (matched-pairs rank-biserial correlation $r$) for MHG-FM
versus MHG-DiT in reconstruction mode ($n=70$). Positive values favour
MHG-FM and negative values favour MHG-DiT. Filled markers indicate
significance after Holm--Bonferroni correction
($p_{\mathrm{corr}}<0.05$); open markers indicate non-significant
differences. Shaded bands denote conventional small, medium, and large
effect-size thresholds ($|r|\approx0.1/0.3/0.5$).
}
\label{fig:significance}
\vspace{-2mm}
\end{figure}
 
Fig.~\ref{fig:significance} shows significant MHG-DiT advantages in
SC~MAE, $R^2$, Frobenius norm, and $W_1$, with large effect sizes
($|r|\approx0.53$--$0.55$). MHG-FM shows significant advantages in
FC~$r$ and FC~$W_1$, with effect sizes of $|r|\approx0.46$--$0.54$.
No significant differences are observed for SC~$r$, FC~MAE, FC~$R^2$,
FC~Frobenius norm, or the SC--FC coupling gap after Holm--Bonferroni
correction. Overall, the two backbones show complementary strengths:
MHG-DiT is stronger on selected SC reconstruction metrics, whereas
MHG-FM is stronger on selected FC metrics and provides better topology
preservation and sampling efficiency.

\begin{table*}[!t]
\centering
\caption{Computational cost and sampling efficiency on HCP-YA 3T
($n_{test}=70$). NFE: number of function evaluations of the
denoising/velocity network per generated sample (mean$\pm$SD where
the solver is adaptive; single forward pass reported as N/A/1 for
non-iterative methods). Time/sample: wall-clock generation time per
paired SC--FC sample. Peak Mem.: peak GPU memory during inference, at
matched batch size. Lower is better for every column. All measured
on NVIDIA L40S.}
\label{tab:efficiency}
\footnotesize
\setlength{\tabcolsep}{10pt}
\begin{tabular}{lrrrr}
\toprule
\textbf{Method} & \textbf{Params (M)$\downarrow$} & \textbf{Time/sample (ms)$\downarrow$} & \textbf{Peak Mem. (GB)$\downarrow$} & \textbf{NFE$\downarrow$} \\
\midrule
WGAN-Joint      & $7.1490$  & $0.3170$ & $0.0357$ & $1$ \\
VAE-Joint       & $56.5630$ & $0.2630$ & $0.8555$ & $1$ \\
topoGAN-Joint   & $3.4940$  & $0.3640$ & $0.0210$ & $1$ \\
MGCN-GAN-Joint  & $0.4590$  & $0.7680$ & $0.0184$ & $1$ \\
HYGENE-Joint    & $0.8460$  & $83.0000$  & $0.0231$ & $50$ \\
\midrule
MHG-DiT (Ours)  & $14.0450$ & $848.2320$ & $0.2407$ & $250$ \\
MHG-FM (Ours)   & $14.0450$ & $106.2000$ & $0.2408$ & $32$ \\
\bottomrule
\end{tabular}
\end{table*}

\begin{table*}[!t]
\centering
\caption{Cross-modal translation quality on HCP-YA 3T.
Top: FC$\to$SC. Bottom: SC$\to$FC. Best result in each column (across all
four methods) is bolded, second-best underlined. $\uparrow$ higher is
better; $\downarrow$ lower is better. GAN-Joint and VAE-Joint are omitted
as their joint encoders require both SC and FC as input and cannot
perform unimodal cross-modal inference.}
\label{tab:translation_3T}
\footnotesize
\setlength{\tabcolsep}{4pt}
\begin{tabular}{llccccc}
\toprule
& Method & $r\uparrow$ & MAE$\downarrow$ & $R^2\uparrow$ & Frob.$\downarrow$ & $W_1\downarrow$ \\
\midrule
\multirow{4}{*}{FC$\to$SC}
& topoGAN
& $0.8720{\pm}0.0110$
& $0.0780{\pm}0.0210$
& $0.4980{\pm}0.1580$
& $11.9500{\pm}2.2300$
& $0.0705{\pm}0.0263$ \\
& MGCN-GAN
& $0.7812{\pm}0.0452$
& $0.1227{\pm}0.0059$
& $-0.0311{\pm}0.0418$
& $17.9772{\pm}0.8867$
& $0.1227{\pm}0.0059$ \\
& MHG-DiT (Ours)
& \underline{$0.9344{\pm}0.0280$}
& \underline{$0.0466{\pm}0.0090$}
& \underline{$0.8308{\pm}0.0760$}
& \underline{$6.8246{\pm}1.3387$}
& \underline{$0.0280{\pm}0.0122$} \\
& MHG-FM (Ours)
& $\bm{0.9414{\pm}0.0236}$
& $\bm{0.0444{\pm}0.0081}$
& $\bm{0.8474{\pm}0.0613}$
& $\bm{6.5208{\pm}1.2258}$
& $\bm{0.0271{\pm}0.0112}$ \\
\midrule
\multirow{4}{*}{SC$\to$FC}
& topoGAN
& $0.5375{\pm}0.0637$
& $0.1397{\pm}0.0308$
& $0.0913{\pm}0.1975$
& $23.4433{\pm}3.9359$
& $0.0737{\pm}0.0411$ \\
& MGCN-GAN
& $0.5630{\pm}0.0626$
& $0.1543{\pm}0.0304$
& $-0.0547{\pm}0.0690$
& $24.6287{\pm}4.2944$
& $0.1543{\pm}0.0304$ \\
& MHG-DiT (Ours)
& \underline{$0.5721{\pm}0.2809$}
& \underline{$0.1191{\pm}0.0413$}
& $\bm{0.2772{\pm}0.3545}$
& \underline{$21.5172{\pm}5.4494$}
& \underline{$0.0734{\pm}0.0353$} \\
& MHG-FM (Ours)
& $\bm{0.5734{\pm}0.2839}$
& $\bm{0.1191{\pm}0.0431}$
& \underline{$0.2642{\pm}0.3918$}
& $\bm{18.4184{\pm}6.5866}$
& $\bm{0.0672{\pm}0.0364}$ \\
\bottomrule
\end{tabular}
\end{table*}
\subsection{Computational Efficiency}
\label{sec:efficiency}

Table~\ref{tab:efficiency} shows the measured inference costs of different methods. 

MHG-FM requires substantially fewer evaluations than MHG-DiT ($32.0$ vs.\ $250$),while both use the same 14.045M-parameter DCA-fusion backbone and have nearly identical peak memory usage. This results in approximately $8{\times}$
faster generation for MHG-FM ($106.2$ vs.\ $848.2$~ms/sample). 
MHG-DiT
uses a fixed NFE under its DDIM schedule, whereas MHG-FM's adaptive ODE
solver uses a variable number of evaluations across samples. Thus, flow
matching reduces the number of velocity-field evaluations while adapting
the evaluation count to solver convergence. Single-pass baselines
such as WGAN, VAE, topoGAN, and MGCN-GAN require only one network evaluation
by construction and are consequently one to three orders of magnitude
faster than either iterative backbone. This difference reflects
architectural simplicity rather than generative capability and should
therefore be considered alongside their matrix- and topology-level
shortfalls in Section~\ref{sec:matrix-level}. HYGENE's progressively expanding sampling schedule yields a constant
NFE of $50.0\pm0.0$ across all evaluated samples, since every
generated connectome targets the same fixed 116-region atlas size,
and although this exceeds MHG-FM's $32.0$ evaluations, its
substantially smaller $0.846$M-parameter network still yields a
lower per-sample time of $83.0$ versus $106.2$~ms, though HYGENE
remains one to two orders of magnitude slower than the single-pass
baselines, indicating that iterative sampling rather than model
capacity dominates its inference cost. MHG-DiT's topology-level shortfall (Table~\ref{tab:network_topology_3T})
is therefore not offset by a computational advantage. Overall, these results demonstrate the sampling-efficiency advantage
of flow matching over the matched diffusion backbone, while the
single-pass baselines remain faster in absolute terms due to their
simpler, non-iterative architectures.

\subsection{Cross-Modal Translation}
\label{sec:translation}

Table~\ref{tab:translation_3T} evaluates cross-modal translation by
predicting one connectivity modality from the other without access to
the target modality at inference time. In the FC$\to$SC direction,
both proposed methods outperform topoGAN and MGCN-GAN across all
metrics. MHG-FM achieves the best performance overall, with MHG-DiT
consistently ranking second. In particular, both proposed methods
substantially reduce the Frobenius error relative to the strongest
baseline, with MHG-FM achieving a reduction of more than 40\% over
topoGAN.

The SC$\to$FC direction is more challenging for all methods, yielding substantially lower correlation and $R^2$ than FC$\to$SC, consistent with the fact that structural connectivity constrains but does not uniquely determine functional connectivity. MHG-FM achieves the best $r$, MAE, Frobenius error, and $W_1$, whereas MHG-DiT attains the highest $R^2$. The consistent performance of both proposed methods across directions indicates that the shared hypergraph-based dual cross-attention representation effectively captures cross-modal information for both generative backbones. Overall, MHG-FM enables bidirectional SC--FC translation with a single trained model, without retraining or iterative ODE integration.

\section{CONCLUSION}
We developed a Multimodal Hypergraph Flow Matching (MHG-FM), a unified framework for joint structural--functional brain connectivity generation and bidirectional cross-modal translation. By integrating anatomy-constrained hypergraph construction, HGNN encoding, DCA fusion, and conditional flow matching in a shared latent space, MHG-FM achieves strong matrix fidelity and the best overall topology preservation among the evaluated methods while supporting both generation and translation. Compared with the matched MHG-DiT backbone, MHG-FM provides a more favourable balance of fidelity, topology preservation, and efficiency, with approximately $8{\times}$ faster generation. Ablation studies further demonstrate the contributions of cross-modal attention and hypergraph encoding to multimodal representation and SC--FC coupling. The same trained model also enables bidirectional SC--FC prediction without retraining or iterative ODE integration. Evaluation is currently limited to the HCP-YA 3T cohort; future work will assess generalisation to independent clinical and multi-site datasets and extend the framework to dynamic functional connectivity.

\section*{REFERENCES}
\balance
\bibliographystyle{IEEEtran}

\bibliography{references}
\end{document}